\documentclass{article}

 \PassOptionsToPackage{numbers, compress}{natbib}

\usepackage[preprint]{neurips_2024}

\usepackage[utf8]{inputenc} 
\usepackage[T1]{fontenc}    
\usepackage[pagebackref=true]{hyperref}       

\usepackage{url}            
\usepackage{booktabs}       
\usepackage{amsfonts}       
\usepackage{nicefrac}       
\usepackage{microtype}      
\usepackage[table]{xcolor}         

\usepackage{algorithm}      
\usepackage{algpseudocode}  
\usepackage{amsmath}        

\definecolor{mycustompurple}{RGB}{154, 36, 79} 
\usepackage[utf8]{inputenc}

\usepackage{hyperref} 

\hypersetup{
    colorlinks=true,            
    linkcolor=red,              
    citecolor=green,            
    filecolor=mycustompurple,   
    urlcolor=magenta            
}

\usepackage{graphicx}
\usepackage{caption}
\usepackage{subcaption}

\usepackage{multirow}
\usepackage{array}
\usepackage{pifont}
\usepackage{float}

\usepackage{wrapfig}
\usepackage{lipsum} 

\usepackage{amssymb}

\title{Exploring Graph-based Knowledge: Multi-Level Feature Distillation via Channels Relational Graph}

\author{%
  Zhiwei Wang$^{1,3}$\quad Jun Huang$^{1}$\quad Longhua Ma$^{2}$\thanks{Correspondence to: Longhua Ma <lhma\_zju@zju.edu.cn>}\quad Chengyu Wu$^{2}$ \quad  Hongyu Ma
  \\
  $^{1}$Ningbo Innovation Center, Zhejiang University, China\quad $^{2}$NingboTech University\quad \\
   \texttt{$^{3}$22360413@zju.edu.cn}\\
}

\begin{document}

\maketitle

\begin{abstract}

  In visual tasks, large teacher models capture essential features and deep information, enhancing performance. However, distilling this information into smaller student models often leads to performance loss due to structural differences and capacity limitations. To tackle this, we propose a distillation framework based on graph knowledge, including a multi-level feature alignment strategy and an attention-guided mechanism to provide a targeted learning trajectory for the student model. We emphasize spectral embedding (SE) as a key technique in our distillation process, which merges the student's feature space with the relational knowledge and structural complexities similar to the teacher network. This method captures the teacher's understanding in a graph-based representation, enabling the student model to more accurately mimic the complex structural dependencies present in the teacher model. Compared to methods that focus only on specific distillation areas, our strategy not only considers key features within the teacher model but also endeavors to capture the relationships and interactions among feature sets, encoding these complex pieces of information into a graph structure to understand and utilize the dynamic relationships among these pieces of information from a global perspective. Experiments show that our method outperforms previous feature distillation methods on the CIFAR-100, MS-COCO, and Pascal VOC datasets, proving its efficiency and applicability. 

\end{abstract}

\section{Introduction}

In the field of deep learning, the impressive success of large neural networks has come at the cost of increased computational complexity, which poses significant challenges for deployment in resource-constrained environments. While these heavyweight models, often referred to as teacher networks, set state-of-the-art benchmarks on various tasks, their practical applicability is limited by their demanding requirements for memory, processing power, and energy. Knowledge Distillation (KD) \cite{hinton2015distilling} emerges as a promising solution to address this dichotomy by transferring the knowledge from a cumbersome model to a more compact and efficient student networks.

The quintessence of knowledge distillation lies in its ability to encapsulate the representational power of a larger model into a smaller one without incurring a substantial loss in performance. Pioneered by Hinton et al. \cite{hinton2015distilling}, the process involves training a smaller model to mimic the behavior of the pre-trained larger model by softening the output logits \cite{jin2023multi}, thus leveraging the rich information embedded in the output distributions of the teacher network.

Recent advances in KD techniques have extended beyond the mere replication of output distributions. Contemporary works explore the distillation of intermediate representations \cite{adriana2015fitnets,yim2017gift,liu2019knowledge}, attention mechanisms \cite{zagoruyko2016paying,kim2021feature}, and even the inculcation of adversarial robustness from teacher to student \cite{maroto2022benefits,zi2021revisiting,shao2021and}. The underlying hypothesis is that the intermediate layers of a neural network embody a wealth of information about the data manifold that, when transferred effectively, can endow the student with nuanced understanding akin to its teacher.

Large and complex teacher models often capture a wealth of features and deep information, crucial for enhancing task performance \cite{dosovitskiy2020image}. However, attempting to directly distill this rich information into smaller capacity student models often results in suboptimal performance reproduction due to the student model's capacity limitations. The student model may struggle to process the complex information present in the teacher model, leading to ineffective distillation. Moreover, simply mimicking all features of the teacher model overlooks the differences in relationships and structures between the information, particularly when there is a significant gap between the teacher and student models \cite{gou2021knowledge}. Recently, researchers have adopted a strategy of carefully selecting distillation areas, aiming to identify and choose key areas that positively impact student learning \cite{chen2017learning,dai2021general,wang2019distilling,li2017mimicking,sun2020distilling,guo2021distilling}. This approach is based on the notion that not all information in the teacher model is equally important for the student model's learning. By focusing on features and areas most beneficial for the student model's performance improvement, more efficient and targeted knowledge transfer can be achieved. Although this selective distillation strategy has seen some success, the process of determining which areas are "key" can be influenced by subjective judgment, leading to different outcomes based on different strategies and criteria. The effectiveness of this method may be limited by the lack of a clear and consistent standard. Over-focusing on local features while ignoring the interdependencies and global structural information among features limits the student model's comprehensive understanding of the teacher model's deep knowledge structure. The student model may perform poorly when faced with new or changing data.

To overcome these challenges, we present a novel distillation framework that delves into the sophisticated feature hierarchies and attention mechanisms naturally acquired by deep convolutional networks. Central to our approach is a multi-level feature alignment strategy that harmonizes the representational layers of the student model with those of the teacher model, moving past traditional feature mimicry. Additionally, we introduce an attention-guided mechanism that highlights the teacher's critical focus areas, enabling a more targeted and efficient learning trajectory for the student model.Inspired by the success of spectral embedding in the field of clustering \cite{luo2003spectral}, we particularly emphasize the introduction of spectral embedding as a cornerstone of our distillation methodology. This advanced technique underpins a nuanced channel-wise distillation process. Recognizing that the channels of the teacher network do not exist in isolation but are interdependent in expressing input data, together forming a complex network of feature interactions \cite{liu2021exploring}, spectral embedding infuses the student model's feature space with the relational knowledge and structural complexities akin to those in the teacher network. Through this method, we aim to enrich the student model with a deeper level of distilled insights, capturing the essence of the teacher's understanding in a graph-based representation. We hope that the use of spectral embedding and graph-based knowledge encapsulation can enrich the learning experience for students and provide a novel perspective to conventional distillation practices. Unlike approaches that target limited distillation regions, our methodology not only identifies crucial features in the teacher model but also aims to grasp the intricate interplay between these features. By representing this sophisticated data as a graph, we can holistically comprehend and leverage the evolving interconnections between various elements, offering a comprehensive view of the knowledge to be distilled.

We conducted comprehensive experiments across a wide spectrum of benchmark datasets, encompassing CIFAR-100 \cite{krizhevsky2009learning} for image classification, MS-COCO \cite{lin2014microsoft} and Pascal VOC \cite{everingham2010pascal} for object detection. Detailed experimental details can be found in Section III. This broad experimental setup allowed us to compare various teacher-student architectures and adaptation layers across multiple visual tasks.

In summary, our contributions in this paper are threefold:
\begin{itemize}
\item We pioneer the modeling of inter-channel correlations within the teacher network as a graphical structure. This methodology lays the foundation for our Spectral Embedding Knowledge Distillation (SEKD) technique.

\item Our method employs spectral embedding to encapsulate the knowledge of the teacher network, and through an attention-guided mechanism, focuses on a holistic understanding across channel-level, relational, and structural dimensions, ensuring the student network inherits a comprehensive representation of the teacher's knowledge.

\item We provide comprehensive empirical evidence that our approach outperforms existing distillation methods, achieving state-of-the-art performance on multiple benchmarks.
\end{itemize}

\section{Method}
\label{Method}



\begin{figure}
    \centering
    \includegraphics[width=1\linewidth]{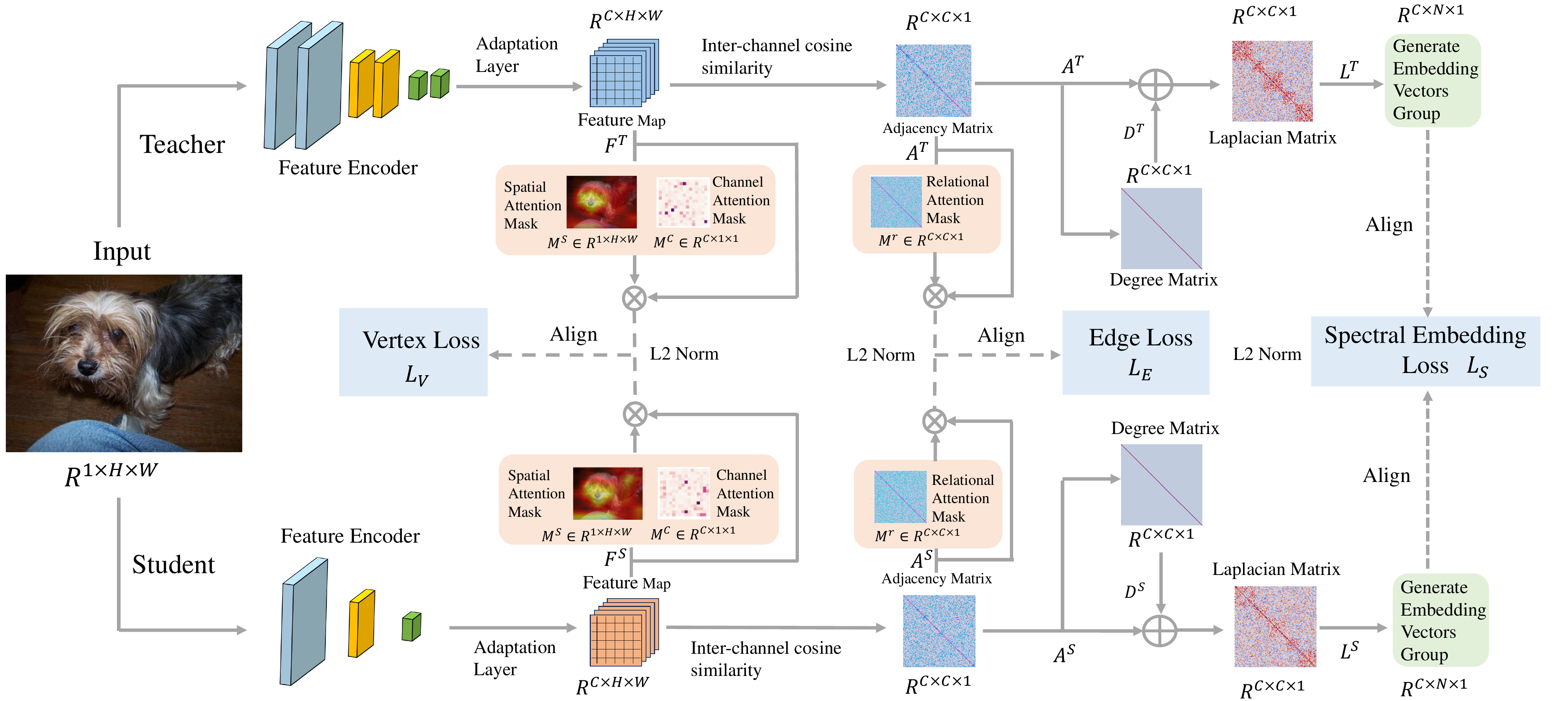}
    \caption{The overall framework of the Multi-Level Feature Distillation. The process of generating embedding vectors in groups is shown in Figure~\ref{figure 2}. We align the channel relational graphs of the teacher and student across multiple levels: vertices, edges, and spectral embeddings.}
    \label{figure 1}
\end{figure}

The overall framework of our method is illustrated in Fig.~\ref{figure 1}. In this section, we will first construct the channels and their relationships of the feature maps into a graph called $\boldsymbol{CRG}$. Then the vertex and edge loss with attention mechanism of $\boldsymbol{CRG}$ will be formulated to constitute the $\boldsymbol{CRG}$ loss. Subsequently, the spectral embedding process and the loss will be described in detail. Finally, we combine the multi-level feature loss to supervise the distillation.

\subsection{Channels Relational Graph}

We use $\boldsymbol{F}\in R^{C\times W \times H}$ to denote the feature maps from one layer of the backbone, where $C$, $W$, $H$ denotes its channel number, width and height. Consider each channel as the vertex and the relation between the channels as the edge. Then the Channels Relation Graph ($\boldsymbol{CRG}$) can be formulated as:

\begin{equation}\label{1}
\boldsymbol{CRG}=(\mathcal{V},\mathcal{E})=(\{\boldsymbol{F}_{k,:,:}\}^C_{k=1},\boldsymbol{A})
\end{equation}

where $\mathcal{V}$ is the vertex set of $\boldsymbol{CRG}$ composed of various channels of feature maps,
$\mathcal{E}$ is the edge set of $\boldsymbol{CRG}$ represented by a weighted adjacency matrix $\boldsymbol{A}\in R^{C\times C}$, the subscript $k,:,:$ of $\boldsymbol{F}_{k,:,:}$ indicates the $k$-th channel of the feature maps in the layer. Each element of the adjacency matrix is obtained by calculating cosine similarity using two flattened feature maps as vectors, which can be formulated as:\\

\begin{equation}\label{2}
\boldsymbol{A}_{ij}=cos\angle vec(\boldsymbol{F}_{i,:,:})vec(\boldsymbol{F}_{j,:,:})=\frac{<vec(\boldsymbol{F}_{i,:,:}),vec(\boldsymbol{F}_{j,:,:})>}{\parallel vec(\boldsymbol{F}_{i,:,:})\parallel \parallel vec(\boldsymbol{F}_{j,:,:})\parallel} ,i,j=1,2,3,...,C,
\end{equation}

where the operator $vec(\cdot)$ vectorize a channel feature map. The adjacency matrix normalizes the pairwise relationship between channels using cosine similarity.

\subsection{Loss for CRG}

\begin{figure}
    \centering
    \includegraphics[width=1\linewidth]{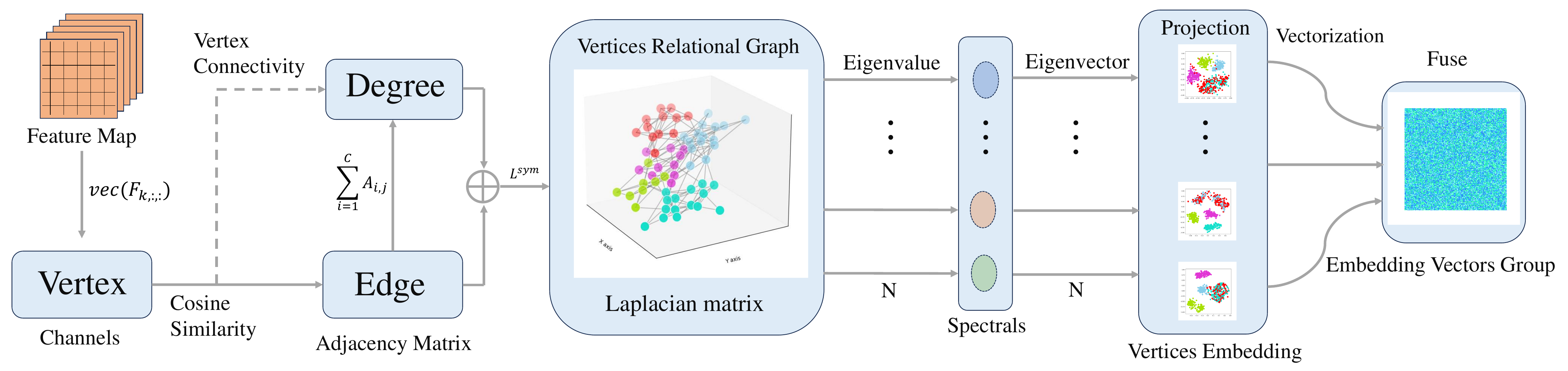}
    \caption{Illustration of generating embedding vectors group. We consider the Laplacian matrix as an algebraic representation of the graph. These embedding vectors provide a structural mapping and topological characteristics of the graph in a low-dimensional subspace.}
    \label{figure 2}

\end{figure}

The $\boldsymbol{CRG}$ Distillation loss function $\mathcal{L}_{CRG}$ consists of two parts: the Vertex loss $L_V$ and the Edge loss $L_E$. Instead of having the student directly mimic the teacher, we utilize the attention mask matrix to guide the distillation. Previous studies have shown that spatial and channel attention masks from teacher models can guide students to focus on learning important foreground information. By overlaying the absolute value of feature maps from teacher model and using softmax function to convert, the attention mask matrices can be formulated as:

\begin{equation}
    \boldsymbol{M}^s=H\cdot W\cdot softmax(\frac{1}{C}\sum_{k=1}^{C}\left|\boldsymbol{F}_{k,:,:}\right|),\:\boldsymbol{M}^{c}=C\cdot softmax(\frac{1}{HW}\sum_{i=1}^{W}\sum_{j=1}^{H}\left|\boldsymbol{F}_{:,i,j}\right|)
\end{equation}
\begin{equation}
   \boldsymbol{M}^{r}=softmax(\left|A\right|)
\end{equation}



where $\boldsymbol{F}_{k,:,:}$ and $\boldsymbol{F}_{:,i,j}$ denotes the $k$-th slice channel and the $i$-th row $j$-th column sequence of $\boldsymbol{F}$ respectively. Note that the superscripts "s", "c" and "r" represent "spatial", "channel" and "relation" respectively. The spatial attention mask matrix $\boldsymbol{M}^s$ implies the important region of the feature map while the channel attention mask matrix $\boldsymbol{M}^c$ implies the important channel in a group of channels. Similarly, the relational attention mask matrix indicates the relationships between the most noteworthy feature maps. $\mathcal{L}_V$ is utilized to enforce the student to absorb the feature maps knowledge by a $L_2$ norm loss under the decoration of $\boldsymbol{M}^s$ and $\boldsymbol{M}^c$, which can be formulated as:

\begin{equation}
    \mathcal{L}_{V}=\frac{1}{CHW}\sum_{k=1}^{C}\sum_{i=1}^{H}\sum_{j=1}^{W}(\boldsymbol{F}_{k,i,j}^{T}-\boldsymbol{F}_{k,i,j}^{S})^2\cdot \boldsymbol{M}^{s}_{i,j}\cdot  \boldsymbol{M}^{c}_{k}
\end{equation}

where the superscripts T and S is utilized to discriminate the teacher model and student model. Likewise, $\mathcal{L}_E$ is utilized to enforce the student to absorb the channel relation knowledge by a $L_2$ norm loss under the decoration of $M^r$, which can be formulated as:

\begin{equation}
    \mathcal{L}_{E}=\frac{1}{C^2}\sum_{i=1}^{W}\sum_{j=1}^{H}(\boldsymbol{A}^T_{i,j}-\boldsymbol{A}^S_{i,j})^2\cdot \boldsymbol{M}^{r}_{i,j}
\end{equation}
\begin{algorithm}
\caption{Multi-Level Feature Distillation}
\label{algorithm:Framwork}  
\begin{algorithmic}[1]
\State $F^S, F^T$: feature maps of student and teacher from layers 1 to L
\State $L_{Total} \gets 0$ \Comment{Initialize the total loss}
\For{$i \gets 1$ \textbf{to} $L$}
    \State $F^{S,v}_i \gets F^S[i].view(F^S[i].size(0), -1)$ \Comment{Vectorize and Normalize student features}
    \State $F^{T,v}_i \gets F^T[i].view(F^T[i].size(0), -1)$ \Comment{Vectorize and Normalize teacher features}
    \State $A^S \gets F^{S,v}_i @ F^{S,v}_i.T$ \Comment{Adjacency matrix for student}
    \State $A^T \gets F^{T,v}_i @ F^{T,v}_i.T$ \Comment{Adjacency matrix for teacher}
    \State $L_V \gets $ \Call{ComputeVertexLoss}{$A^S, A^T$} \Comment{Vertex loss based on feature similarity}
    \State $L_E \gets $ \Call{ComputeEdgeLoss}{$A^S, A^T$} \Comment{Edge loss from graph theory perspective}
    \State $D^S \gets torch.diag(A^S.sum(1))$
    \State $D^T \gets torch.diag(A^T.sum(1))$
    \State $L^{sym,S} \gets torch.eye(D^S.size(0)) - D_{S}^{-0.5} @ A^S @ D_{S}^{-0.5}$ \Comment{Laplacian for student}
    \State $L^{sym,T} \gets torch.eye(D^T.size(0)) - D_{T}^{-0.5} @ A^T @ D_{T}^{-0.5}$ \Comment{Laplacian for teacher}
    \State $E^S, \_ \gets torch.linalg.eigh(L^{sym,S})$
    \State $E^T, \_ \gets torch.linalg.eigh(L^{sym,T})$
    \State $L_S \gets $ \Call{ComputeSpectralLoss}{$E^S, E^T$} \Comment{Spectral embedding loss from eigenvalues}
    \State $L_M \gets L_V + L_E + L_S$ \Comment{Combine losses for multiple distillation levels}
    \State $L_{Total} \gets L_{Total} + L_M$ \Comment{Accumulate total loss across all layers}
\EndFor
\State $L_{Total} \gets L_{Total} + L_{Original}$ \Comment{Include any original task-specific loss}
\end{algorithmic}
\end{algorithm}

\subsection{Loss for Spectral Embedding}

Spectral Embedding is an effective method to extract the features of a group of data points by reducing the dimension and retaining the key information, which is often used in clustering algorithms. In our method, we apply the spectral embedding to capture the geometric and manifold structure of the $\boldsymbol{CRG}$. In order to perform spectral embedding processing on $\boldsymbol{CRG}$, we need to calculate the weighted degree matrix $\boldsymbol{D}$ and the the symmetric normalized Laplacian matrix $L^{sym}$, which can be formulated as:

\begin{equation}
    \boldsymbol{D}_{i,i}=\sum_{i=1}^{C}\boldsymbol{A}_{i,j}, L^{sym}=I-\boldsymbol{D}^{-1/2}\boldsymbol{AD}^{-1/2}
\end{equation}


where the weighted degree matrix $\boldsymbol{D}$ is a diagonal matrix and the elements on the diagonal $\boldsymbol{D}_{i,i}$ is calculated by summing the $i$-th column of the adjacency matrix $\boldsymbol{A}$. Note that $I$ is an identity matrix. Due to $L^{sym}$ being a symmetric matrix, we can perform eigenvalue decomposition on it as $L^{sym}=\boldsymbol{U}\Lambda \boldsymbol{U}^T$.
Where $\Lambda$ is a diagonal matrix composed of the $C$ eigenvalues of a diagonal element $L^{sym}$ and $\boldsymbol{U}$ is an eigenvector matrix composed of linearly independent eigenvectors corresponding to $C$ eigenvalues. By selecting the eigenvectors corresponding to the top $N$ largest eigenvalues, we can form a spectral embedding vectors group matrix $\boldsymbol{E}$. The hyperparameter sensitivity experiment for $N$ is detailed in the Appendix \ref{N}. We provide a detailed explanation of the process of generating spectral embeddings group in Fig. \ref{figure 2}. Now, we can define the spectral embedding loss function with a $L_2$ norm loss of $\boldsymbol{E}$ to enforce the student to imitate the teacher model in a reduced dimensional space, which can be formulated as:

\begin{equation}
   \mathcal{L}_{S}=\frac{1}{CN}\sum_{i=1}^{C}\sum_{j=1}^{N}(\boldsymbol{E}^T_{i,j}-\boldsymbol{E}^S_{i,j})^2
\end{equation}

\subsection{Multi-Level Feature Loss And Overall Loss}

Now we have designed the Vertex Loss, the Edge loss and the Spectral Embedding loss, and by introducing three hyper-parameters to balance them the Multi-Level Feature Loss can be combined as:

\begin{equation}
    \mathcal{L}_{M}=\alpha\cdot \mathcal{L}_{V}+\beta\cdot \mathcal{L}_{E}+\gamma\cdot \mathcal{L}_{S},\: \mathcal{L}_{Total}={L}_{M}+{L}_{Origin}
\end{equation}


Wherein, ${L}_{Origin}$ represents the loss function between the prediction results and the Ground Truth. ${L}_{M}$ denotes the Multi-Level Feature Loss. ${L}_{Total}$ denotes the global loss function. Our source of knowledge comes from the feature representations of the intermediate layers. Therefore, it can be generalized to various detectors, various teacher-student architectures, and other visual tasks.

Now that we have introduced the calculation methods for all loss functions, we can select several important layers in the teacher and student networks for distillation. The main flow of the Multi-Level Feature Distillation is summarized in Algorithm \ref{algorithm:Framwork}. For the sensitivity experiments pertaining to the hyperparameters $\alpha$, $\beta$ and $\gamma$, please refer to Section \ref{Hyperparameter sensitivity.}.

\begin{table}[htbp]

\caption{We benchmark our approach by comparing it with other state-of-the-art (\textbf{SOTA}) methods across various detectors. In this comparison, the teacher network employed is ResNet101, while the student network is ResNet50. The evaluation is conducted on the MS-COCO val2017 dataset. \textsuperscript{\textdagger} indicates that the method was reproduced by us, while the remaining results are sourced from the related paper.}
\label{table 1}
\begin{minipage}{0.5\linewidth}
\centering
\resizebox{\linewidth}{!}{
\begin{tabular}{p{6em}lcccccc}
\hline
Method & AP & AP\textsubscript{50} & AP\textsubscript{75}& AP\textsubscript{S} & AP\textsubscript{M} & AP\textsubscript{L} \\
\hline
\multicolumn{7}{c}{\textbf{Faster RCNN \cite{faster2015towards}-FPN \cite{lin2017feature}}} \\ 
\textbf{Teacher}  &42.0 & 62.5  &45.9   &25.2  &45.6   & 54.6  \\
\textbf{Student}  & 37.4  & 58.4  & 40.7  & 21.8  &41.0   & 47.8   \\
\textsuperscript{\textdagger}KD \cite{hinton2015distilling}   &38.3   &59.4   & 41.7  &22.2   &41.9   &50.4\\
\textsuperscript{\textdagger}FitNet \cite{adriana2015fitnets}  &38.8  &59.6  &41.8   &22.3   &42.2   &50.7\\
FGFI \cite{wang2019distilling}  & 39.4  & 60.3  & 43.0  &22.9   & 43.6 &52.8  \\
ReviewKD \cite{chen2021distilling}    &40.4   &61.0   &44.0   & 23.6  &43.8  &52.9  \\
MKD \cite{jin2023multi}    &40.2   &61.6   &44.5   & 23.2  &44.5  &53.4  \\
\rowcolor{gray!25}
\textbf{Ours}    &\textbf{41.9}  & \textbf{61.8}  &\textbf{45.5}  &\textbf{24.1}   &\textbf{44.6}  &\textbf{54.0}  \\
\hline
\multicolumn{7}{c}{\textbf{RetinaNet \cite{lin2017focal}}} \\ 
\textbf{Teacher} & 38.9 & 58.0 & 41.5 & 21.0 & 42.8 & 52.4 \\
\textbf{Student} & 37.4 & 56.7 & 39.6 & 20.0 & 40.7 & 49.7 \\
\textsuperscript{\textdagger}FitNet \cite{adriana2015fitnets}  & 37.4 & 57.1 & 40.0 & 20.8 & 40.8 & 50.9 \\
FGFI \cite{wang2019distilling} & 39.0 & 58.8  & 41.9 & 21.7 & 42.9 &52.2  \\
FRS \cite{zhixing2021distilling} & 39.3 & 58.8 & 42.0 & 21.5 & 43.3 & 52.6 \\
GI \cite{dai2021general} & 39.1 & 59.0 & 42.3 & 22.8 & 43.1 & 52.3  \\
\textsuperscript{\textdagger}FGD \cite{yang2022focal} & 39.6 & 59.4 & 42.7 & 22.9 & 43.7 & 53.6 \\
\rowcolor{gray!25}
\textbf{Ours}    &\textbf{39.9} &\textbf{59.8}  &\textbf{43.2}  &\textbf{23.6}   &\textbf{44.1}  &\textbf{54.0} \\
\hline
\end{tabular}
}
\end{minipage}%
\begin{minipage}{0.5\linewidth}
\centering
\resizebox{\linewidth}{!}{
\begin{tabular}{p{6em}lcccccc}
\hline
Method & AP & AP\textsubscript{50} & AP\textsubscript{75}& AP\textsubscript{S} & AP\textsubscript{M} & AP\textsubscript{L} \\
\hline
\multicolumn{7}{c}{\textbf{FCOS \cite{tian2022fully}}} \\
\textbf{Teacher} & 40.8  & 60.0  & 44.0  & 24.2  &44.3   & 52.4  \\
\textbf{Student}  &38.5 & 57.7  &41.0   &21.9  &42.8   & 48.6    \\
\textsuperscript{\textdagger}FitNet \cite{adriana2015fitnets}   &39.9   &58.6   &43.1  &23.1   &43.4  &52.2\\
GI \cite{dai2021general}  & 42.0  & 60.4  & 45.5  &25.6   & 45.8 &54.2  \\
\textsuperscript{\textdagger}FGD \cite{yang2022focal}   &42.1  &60.6   &45.9   & 27.0  &46.0  &54.6  \\
FRS \cite{zhixing2021distilling}    &40.9   &60.3   &43.6  & 25.7  &45.2  &51.2  \\
MasKD \cite{huang2022masked}    &42.6   &61.2   &46.3   & 26.5  &46.9  &54.2  \\
\rowcolor{gray!25}
\textbf{Ours}     &\textbf{43.1}  &\textbf{61.4}   &\textbf{46.5}   &\textbf{27.3}   &\textbf{47.9}  &\textbf{55.6}  \\
\hline
\multicolumn{7}{c}{\textbf{GFL \cite{li2020generalized}}} \\
\textbf{Teacher} & 45.0 & 63.7 & 48.9 &27.2& 48.8 & 54.5 \\
\textbf{Student} & 40.1 & 58.2 & 43.1 &23.3& 44.4 & 52.5 \\
\textsuperscript{\textdagger}FitNet \cite{adriana2015fitnets} & 40.7 & 58.6 & 44.0&23.7 & 44.4 & 53.2 \\
GI \cite{dai2021general} & 41.5 & 59.6 & 45.2 &24.3& 45.7 & 53.6 \\
\textsuperscript{\textdagger}FGD \cite{yang2022focal}  &41.6  &59.3   &44.8   &24.5   &45.7  &53.7  \\
PGD \cite{wang2022head}  & 41.8  & 59.8  & 45.2  & 25.0  & 46.4 & 53.4 \\
LD \cite{zheng2023localization} & 42.1 & 60.3 & 45.6  & 24.5 & 46.2 & 54.8 \\
\rowcolor{gray!25}
\textbf{Ours}    & \textbf{42.7}  & \textbf{60.9}   & \textbf{46.2}   &\textbf{25.0}   &\textbf{46.3}  &\textbf{55.1}  \\
\hline
\end{tabular}
}
\end{minipage}
\end{table}

\section{Experiments}
\label{Experiments}

Details of the experimental implementation can be found in Appendix \ref{Experiments Setting}. 

\subsection{Main Results}

\textbf{Comparison to State-of-The-Arts on Ms-COCO.}
In spectral graph theory, spectral embedding captures the global structure and relationships of data by encoding implicit relationships between features, including group clustering and connectivity. Our approach integrates multi-level joint distillation of graph-based vertices, edges, and spectral embedding, aiming to enable the student model to transcend limitations of model capacity and structural differences, thus fully understanding and accurately replicating the behavior of the teacher model. In our research on object detection tasks on the MS-COCO dataset, our method demonstrates consistent performance across various detection architectures. As shown in Table~\ref{table 1}, our training strategy outperforms the simple feature mimicking of FitNet \cite{adriana2015fitnets} by 7.9\% AP in Faster RCNN \cite{faster2015towards}-FPN \cite{lin2017feature}, and exceeds the focus distillation strategies of FGFI \cite{wang2019distilling}, FRS \cite{zhixing2021distilling}, GI \cite{dai2021general}, and FGD \cite{yang2022focal} by 0.9, 0.6, 0.8, and 0.3 AP respectively in RetinaNet \cite{lin2017focal}. By focusing on the structural relationships between foreground and background areas, our method surpasses the high-precision localization strategy LD \cite{zheng2023localization} by 1.4 $\text{AP}_{75}$ in GFL \cite{li2020generalized}. MasKD \cite{huang2022masked} employs masking to extract pixel-level relational embeddings from feature maps, yet its performance on FCOS \cite{tian2022fully} still falls 0.5 AP short of our method.

\textbf{Experimental results on Pascal VOC are summarized in the Appendix \ref{PASCAL VOC}}

\textbf{Various student backbones.}
As shown in Table \ref{table 2}, we opted for the compact one-stage detectors GFL \cite{li2020generalized} as the embeddable frameworks, utilizing ResNet101 as the teacher backbone network. We reported the competitive advantage of our method in enhancing the performance of lightweight student networks such as ResNet34, ResNet18, and MobileNetv2, in comparison to CM \cite{yang2023context}. Notably, our proposed technique consistently improved the student networks' AP by approximately 4\%. Under heterogeneous teacher-student settings (ResNet101 to MobileNetv2), the improvements in $\text{AP}_{S}$, $\text{AP}_{M}$, and $\text{AP}_{L}$ relative to CM \cite{yang2023context} are around 6\%, 5\%, and 2\%, respectively.

\begin{table}[ht]
\caption{Our algorithm has been evaluated under various teacher-student architectural configurations, including heterogeneous MobileNetv2 and homogenous ResNet34 and ResNet18 architectures. The teacher model utilized throughout the assessments is ResNet101, with the MS-COCO val2017 dataset serving as the testing benchmark.}
\label{table 2}
\centering
\begin{tabular}{l|lcccccc}
\hline
Backbone    & Method       & AP   & AP$_{50}$ & AP$_{75}$ & AP$_{S}$  & AP$_{M}$  & AP$_{L}$  \\ \hline
 & Student      & 34.5 & 51.3 & 37.1 & 18.9 & 37.6 & 45.4 \\
MobileNetV2 & CM \cite{yang2023context}         & 35.7 & 52.3 & 38.8 & 19.5 & 38.9 & 47.1 \\ 
             &\textbf{Ours}  \cellcolor{gray!25} &\textbf{37.4} \cellcolor{gray!25}   & \textbf{55.9} \cellcolor{gray!25}  & \textbf{40.1} \cellcolor{gray!25}   & \textbf{22.4} \cellcolor{gray!25}  &\textbf{40.4} \cellcolor{gray!25}  &\textbf{48.5} \cellcolor{gray!25}\\
             \hline
    & Student      & 35.5 & 52.7 & 37.9 & 18.9 & 38.6 & 46.8 \\
ResNet18            & CM \cite{yang2023context}         & 37.6 & 54.6 & 40.6 & 20.3 & 41.0 & 49.7 \\ 
            &\textbf{Ours}  \cellcolor{gray!25}& \textbf{38.4} \cellcolor{gray!25}  & \textbf{57.1} \cellcolor{gray!25}  &  \textbf{42.3} \cellcolor{gray!25}  & \textbf{22.9} \cellcolor{gray!25}  & \textbf{41.6} \cellcolor{gray!25} &\textbf{51.2} \cellcolor{gray!25}\\
            \hline
    & Student      & 39.1 & 56.7 & 42.3 & 21.5 & 43.1 & 51.9 \\
ResNet34   & CM \cite{yang2023context}  & 40.8 & 58.4 & 44.3 & 22.8 & 44.9 & 53.8 \\ 
            &\textbf{Ours}  \cellcolor{gray!25}& \textbf{41.3} \cellcolor{gray!25}&\textbf{59.8 }\cellcolor{gray!25} & \textbf{45.6} \cellcolor{gray!25}&\textbf{24.2} \cellcolor{gray!25}&\textbf{45.1} \cellcolor{gray!25}&\textbf{54.6} \cellcolor{gray!25}\\
            \hline
    & Student      & 40.1 & 58.2 & 43.1 & 23.3 & 44.4 & 52.5 \\
ResNet50           & CM \cite{yang2023context}         & 42.0 & 59.7 & 45.7 & 24.2 & 46.3 & 54.4 \\ 
            &\textbf{Ours}   \cellcolor{gray!25}&\textbf{42.7} \cellcolor{gray!25}& \textbf{60.9} \cellcolor{gray!25}& \textbf{46.2} \cellcolor{gray!25}&\textbf{25.0} \cellcolor{gray!25}&\textbf{46.3} \cellcolor{gray!25}&\textbf{55.1} \cellcolor{gray!25}\\
            \hline
\end{tabular}
\end{table}

\begin{figure}[H]
  \centering
  \includegraphics[width=1\linewidth]{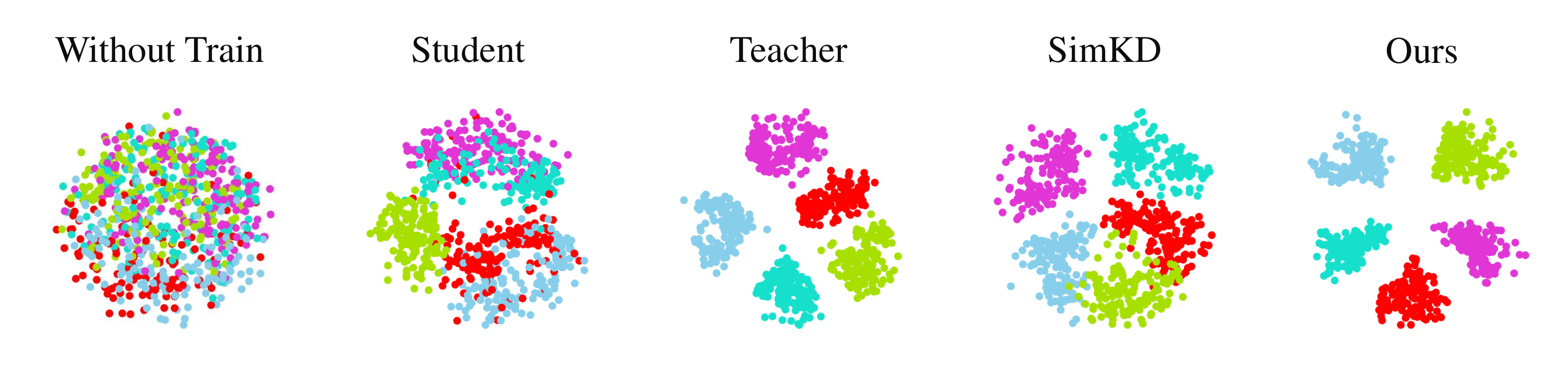}
  \caption{We visualized the classification effects of classifiers using our method and the SimKD \cite{chen2022knowledge} method through t-SNE dimensionality reduction. We also displayed the classification effects of the teacher, student, and untrained models. Multi-level joint alignment based on inter-channel relationships indirectly improves the classification performance between sample points by enhancing the representation of internal channel relationships in the network. We attribute this enhancement to spectral embedding technology, which boosts the student model's understanding of complex data structures, thereby increasing intra-class compactness and inter-class separability in classification.}
    \label{figure 3}
\end{figure}

\textbf{Generalization to classification task.}
Our method demonstrates minimal coupling with object detection tasks, facilitating its seamless extension to other visual tasks like image classification. We assess our method's competitive advantage relative to other approaches in both homogeneous and heterogeneous (Table \ref{table 3}) environments. Notably, despite significant disparities in the topological structures of intermediate layer outputs between teacher and student networks in heterogeneous environments, our method still empowers students to surpass teachers in certain scenarios. For instance, the teacher ResNet-110x2 lags behind the distilled student ShuffleNetV2 by 1.14 in top-1 accuracy. Furthermore, employing the method we introduce with WRN-40-2 results in a top-1 accuracy surpassing that of the teacher ResNet-32x4 by 0.99. As illustrated in Fig.~\ref{figure 3}, we have randomly selected five categories from the CIFAR-100 dataset and visualized the classification features using t-SNE, with each color representing a distinct category. The teacher network is comprised of ResNet110x2, in contrast to the student network which utilizes ShuffleNetV2.
\begin{table}[ht]
\caption{Performance in classification task on the CIFAR-100 dataset. Top-1 accuracy is adopted as the evaluation metric.}
\label{table 3}
\centering
\resizebox{\textwidth}{!}{
\begin{tabular}{l|ccc|ccc}
\hline
Method&\multicolumn{3}{c|}{Homogeneous architecture}&\multicolumn{3}{c}{Heterogeneous architecture}\\
\hline
\multirow{2}{*}{Teacher}& WRN-40-2&ResNet-32x4 &VGG13 &ResNet-110x2&WRN-40-2&ResNet-32x4\\
       &76.31&79.42   &74.64 &78.18&76.31&79.42\\
\hline
\multirow{2}{*}{Student}& WRN-40-1 & ResNet-8x4  & VGG-8& ShuffleNetV2 &MobileNetV2&WRN-40-2 \\
       & 71.92  & 73.09     & 70.46 &72.60 &65.43 &76.35   \\
\hline
KD \cite{hinton2015distilling} & 74.12  & 74.42      & 72.98   &76.05   &69.07&77.70\\
FitNet \cite{adriana2015fitnets}& 74.17  & 74.32      & 71.02  &76.40  &68.64 &77.69  \\
AT \cite{zagoruyko2016paying} & 74.67  & 75.07      & 71.90   &76.84 &68.62 &78.45 \\
SP \cite{tung2019similarity}  & 73.90  & 74.29     & 73.12  &76.60  &68.73 &78.34\\
CRD \cite{tian2019contrastive} & 74.80  & 75.59      & 73.94  &76.67  &70.28  &78.15\\
SimKD \cite{chen2022knowledge} & 75.56  & 78.08     &73.76 &78.25 &70.71&79.29\\
\rowcolor{gray!25}
\textbf{Ours}  &\textbf{75.99}&\textbf{78.89}&\textbf{74.57}&\textbf{79.32}&\textbf{71.18}&\textbf{80.41}\\
\hline
\end{tabular}
}
\end{table}
\subsection{Analyses}

\textbf{Effects of the adaptation layer.}
The role of the projection layer is to align the outputs of the teacher and student at the same semantic level, aiming to reduce additional noise introduced during feature tensor alignment. We compared two types of linear projection: point convolution and 3x3 convolution. The hyperplane projection of the activation layer output and the two types of transfer layers embedded in the network structure, including the regression-based transfer proposed in FitNet \cite{adriana2015fitnets} and the attention-based transfer in AT \cite{zagoruyko2016paying}, were also evaluated. Since the parameters of the convolution layers are randomly assigned, we compared the average AP obtained from 5 experiments. We use Faster RCNN \cite{faster2015towards} with FPN \cite{lin2017feature} as the detector for benchmark testing. The results are shown in Table \ref{table 4}. We found that the noise suppression effect of the 3x3 convolution in our proposed method is significantly better than other types of projection layers.

\begin{table}[H]
\caption{The table examines the influence of different adaptation layers on the performance of a student model ResNet50, with ResNet101 as the teacher model, on the MS-COCO val2017 dataset. Performance metrics are analyzed in relation to various transformer types and their positions within the network architecture. \textsuperscript{\textdagger} describing the strategy used by the original method.}
\label{table 4}
\centering
\resizebox{\textwidth}{!}{
\begin{tabular}{l|lccccc}
\hline
Detector&Method& AP & AP\textsubscript{50} & AP\textsubscript{75}&Transformer&Feature position \\
\hline
\multirow{12}{*}{\text{Faster RCNN \cite{faster2015towards}-FPN \cite{lin2017feature}}}&Teacher & 42.0 & 62.5 & 45.9&-& - \\
&Student & 37.4 & 58.4 & 40.7&-& -\\
&FitNet    &38.9&59.4&41.7&1$\times$1 conv& Mid  layer\\
&FitNet    &38.5&59.7&41.8&3$\times$3 conv& Mid  layer\\
&FitNet \cite{adriana2015fitnets}    &38.8&59.6&41.8&\textsuperscript{\textdagger}regressor& Mid  layer\\
&AB \cite{heo2019knowledge}       &39.1  &59.9 &42.0 &Pre-ReLu& End of group\\
&AT \cite{zagoruyko2016paying}        &39.6  &60.2 &42.1&Attention& End of group\\

&\textbf{Ours}\cellcolor{gray!25}   &41.5\cellcolor{gray!25} &61.3\cellcolor{gray!25}  &45.4\cellcolor{gray!25} &1$\times$1 conv \cellcolor{gray!25}&Mid  layer\cellcolor{gray!25}\\
&\textbf{Ours} \cellcolor{gray!25}   &\textbf{41.9}\cellcolor{gray!25} &\textbf{61.8}\cellcolor{gray!25}  &\textbf{45.5}\cellcolor{gray!25} &3$\times$3 conv \cellcolor{gray!25}&Mid  layer\cellcolor{gray!25}\\
&\textbf{Ours} \cellcolor{gray!25}   &41.6\cellcolor{gray!25} &61.1\cellcolor{gray!25}  &45.0\cellcolor{gray!25} &Regressor \cellcolor{gray!25}&Mid  layer\cellcolor{gray!25}\\
&\textbf{Ours} \cellcolor{gray!25}   &41.3\cellcolor{gray!25} &61.7\cellcolor{gray!25}  &44.9\cellcolor{gray!25} &Pre-ReLu\cellcolor{gray!25}& End of group\cellcolor{gray!25}\\
&\textbf{Ours} \cellcolor{gray!25}   &41.7\cellcolor{gray!25} &61.5\cellcolor{gray!25}  &45.3\cellcolor{gray!25} &Attention\cellcolor{gray!25}& End of group\cellcolor{gray!25}\\
\hline
\end{tabular}
}
\end{table}
\begin{wrapfigure}{r}{0.5\textwidth} 
  \centering
  \includegraphics[width=0.5\textwidth]{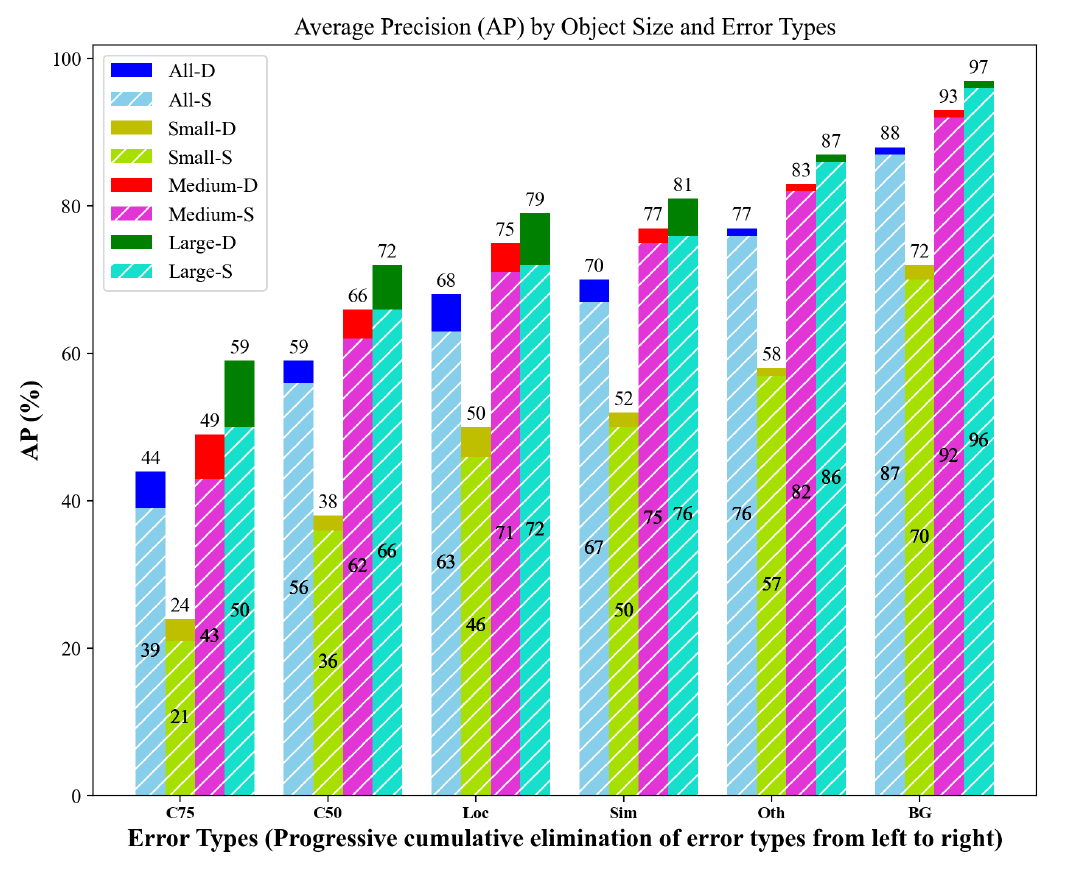}
  \caption{RetinaNet R-50 AP score for various box sizes. S denotes the types of errors made by the student. D represents the types of errors after applying our method.}
  \label{figure 4}
\end{wrapfigure}
\textbf{Error types.} As shown in Fig.~\ref{figure 4}, we reanalyzed the mechanism of the proposed mothod from the perspective of error components. The horizontal axis of the bar chart represents the errors disregarding the corresponding categories, while the vertical axis represents AP. C75 and C50 represent different Intersection over Union (IoU) thresholds, Loc corresponds to $\text{AP}_{10}$, Sim and Oth respectively indicate confusions with similar and other categories, and BG denotes background false positives. Our method enhances the discriminative capabilities of the student, explicitly evidenced by the significant reduction in the proportion of classification errors (Sim and Oth) in False Positives.


\textbf{Combination with response-based distillation Method.}
We investigated the integration of response-based distillation methods into the proposed framework. As shown in Table \ref{table 5}, incorporating $\text{KD}_{cls}$ resulted in a gain of 1.2 $\text{AP}_{S}$. Interestingly, contrary to intuition, the effectiveness of incorporating category regression distillation $\text{KD}_{cls}$ was significantly better than that of incorporating bounding box prediction regression $\text{KD}_{loc}$. We also observed that the fused method's AP was lower than that before fusion, suggesting conflicting learning objectives between intermediate and output layers leading to a decrease in distillation performance.

\begin{table}[H]
\caption{Our method is combined with a response-based distillation approaches. “cls” indicates category prediction, and “loc” corresponds to bounding box prediction. The teacher network is ResNet101, and the student network is ResNet50. The dataset employed for evaluation is MS-COCO.}
\label{table 5}
\centering
\resizebox{\textwidth}{!}{
\begin{tabular}{l|lcccccc}
\hline
Detector&Method & AP & AP$_{50}$  &AP$_{75}$ & AP$_{S}$ & AP$_{M}$ & AP$_{L}$\\
\hline
\multirow{6}{*}{\textbf{Faster RCNN \cite{faster2015towards}-FPN \cite{lin2017feature}}}&Teacher&42.0  &62.5  &45.9&25.2&45.6&54.6\\
&Student &37.4  &58.4  &40.7&21.8&41.0&47.8\\
&\textbf{Ours} \cellcolor{gray!25}&\textbf{41.9}\cellcolor{gray!25}  &61.8\cellcolor{gray!25}  &45.5\cellcolor{gray!25}  &24.1\cellcolor{gray!25}&\textbf{44.6}\cellcolor{gray!25}&\textbf{54.0}\cellcolor{gray!25} \\
&$+$ KD$_{cls}$\cellcolor{gray!25}   &41.8\cellcolor{gray!25}  &61.4\cellcolor{gray!25}  &\textbf{46.2}\cellcolor{gray!25} &\textbf{25.3}\cellcolor{gray!25}&44.1\cellcolor{gray!25}&52.6\cellcolor{gray!25} \\

&$+$ KD$_{loc}$\cellcolor{gray!25}   &41.4\cellcolor{gray!25}  &61.1\cellcolor{gray!25}  &45.3\cellcolor{gray!25} &23.8\cellcolor{gray!25}&44.3\cellcolor{gray!25}&53.3\cellcolor{gray!25} \\

&$+$ KD$_{cls}$ $+$ KD$_{loc}$\cellcolor{gray!25} &41.8\cellcolor{gray!25}   &\textbf{61.9} \cellcolor{gray!25}  &45.2\cellcolor{gray!25}  &24.3\cellcolor{gray!25} &44.5\cellcolor{gray!25} &53.7\cellcolor{gray!25} \\
\hline
\end{tabular}
}
\label{tab:detector_results}
\end{table}

\begin{figure}[H]
    \centering
    \begin{subfigure}[b]{0.32\textwidth}
        \includegraphics[width=\textwidth]{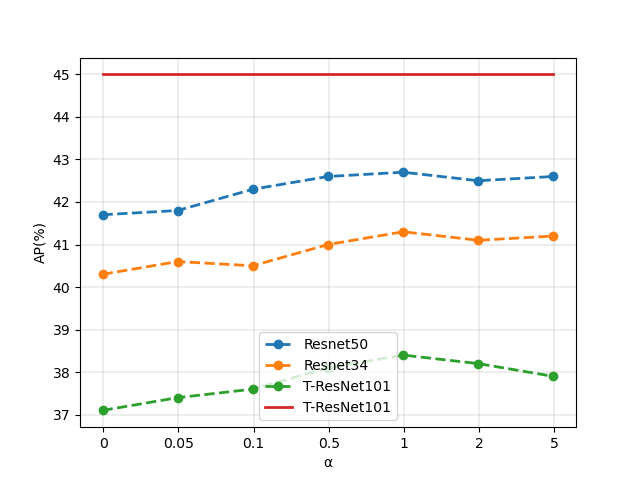}
        \caption{$\alpha$ for vertex loss.}
    \end{subfigure}
    \begin{subfigure}[b]{0.32\textwidth}
        \includegraphics[width=\textwidth]{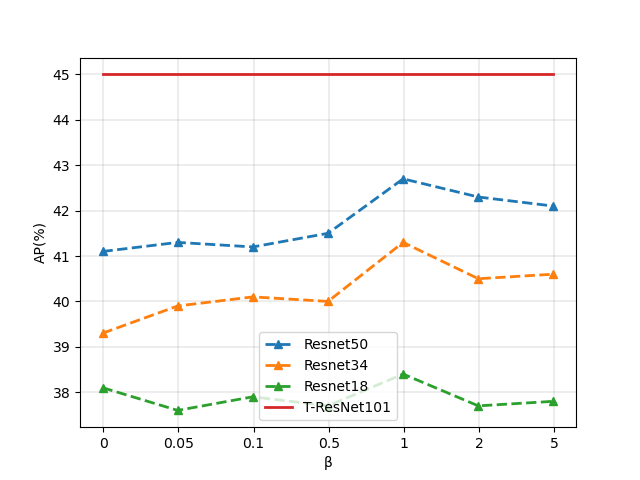}
        \caption{$\beta$ for edge loss.}
    \end{subfigure}
    \begin{subfigure}[b]{0.32\textwidth}
        \includegraphics[width=\textwidth]{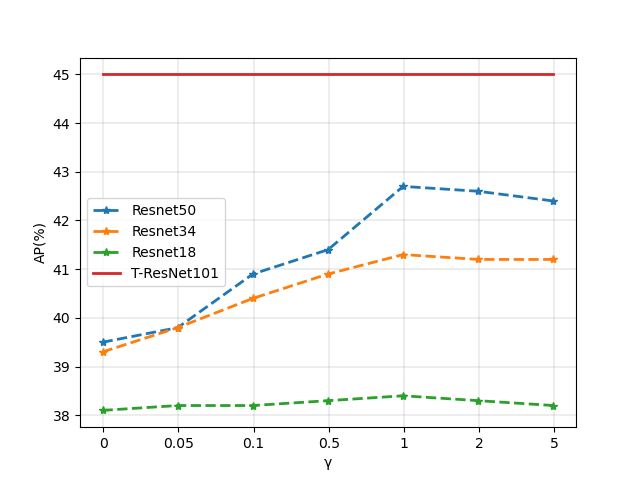}
        \caption{$\gamma$ for SE loss.}
    \end{subfigure}
    \caption{
For each hyperparameter, we observed its impact trends on three different student networks.}
\label{figure 5}
    \label{}
\end{figure}

\textbf{Hyperparameters sensitivity.}\label{Hyperparameter sensitivity.}
As shown in Fig. \ref{figure 5}, we assess the sensitivity of three hyperparameters using GFL \cite{li2020generalized} detector. The model shows insensitivity to adjustments in $\alpha$ but shows a notable change in AP with adjustments to $\gamma$, performing optimally under a relatively balanced hyperparameter setting. We set $\alpha=\beta=\gamma=1$ to ensure the repeatability of the experiments.

\begin{wraptable}{r}{0.5\textwidth} 
\caption{Employing the Faster RCNN \cite{faster2015towards}-FPN \cite{lin2017feature} detector within a ResNet101-50 teacher-student architecture on COCO. L$_{V}$ denotes vertex loss,  L$_{E}$ edge loss, and  L$_{S}$  spectral embedding loss. We incorporate attention masks by default.}
\label{table 6}
\centering
\begin{tabular}{ccc|ccc}
\hline
\multicolumn{3}{c|}{\textbf{Loss}}&\multicolumn{3}{c}{\textbf{Result}}\\
\hline
L$_{V}$ & L$_{E}$ & L$_{S}$ & AP & AP$_{50}$ & AP$_{75}$  \\
\hline
  \ding{55} & \ding{55} & \ding{55} &37.4 &58.4 &40.7  \\
\checkmark & \ding{55} & \ding{55} &38.6 &59.8 &41.9  \\
 \ding{55} & \checkmark & \ding{55} &37.8 &58.9 &43.2  \\
  \ding{55} & \ding{55} & \checkmark &41.5 &61.6 &45.1  \\
   \ding{55} & \checkmark & \checkmark &41.6 &61.4 &45.3  \\
 \checkmark & \checkmark & \ding{55} &39.7 &59.2 &41.4  \\
 \checkmark & \ding{55} & \checkmark &40.3 &61.1 &44.9  \\
 \rowcolor{gray!25}
  \checkmark & \checkmark & \checkmark &\textbf{41.9} &\textbf{61.8} &\textbf{45.5}  \\
\hline
\end{tabular}
\end{wraptable}

\textbf{Ablation study.}  We investigated the contributions of various components in our proposed method to the distillation performance, including vertex loss, edge loss, and spectral embedding loss. As shown in Table \ref{table 6}, we discovered that the combination of edges and SE (Spectral Embedding) vertices performs better than distilling vertices alone, indicating that various features exhibit chaotic relationships. The combination of vertices and edges is less effective than the combination of vertices and SE, suggesting that the student model may struggle to establish connections between the spatial topology of local features and the overall topology through surface mimicry alone. Employing all four components simultaneously yields better results than other variants, demonstrating that these components are orthogonal to each other and each is indispensable.\textbf{
We conducted ablation studies on two types of attention masks, with experimental results  in the Appendix \ref{Attention Mechanism Ablation}}.

\section{Conclusion}

In summary, We innovatively apply spectral embedding (SE) technology to precisely extract structural knowledge from channels relational graph. By using attention masks to guide the vertices and edges of the relational graph, and combining this with spectral embedding, we achieve multi-level fine alignment. Our framework not only surpasses traditional feature distillation methods in efficiency but also achieves unparalleled performance across benchmark datasets.

\bibliographystyle{plain}
\bibliography{References}

\begin{thebibliography}{10}

\bibitem{adriana2015fitnets}
Romero Adriana, Ballas Nicolas, K~Samira Ebrahimi, Chassang Antoine, Gatta Carlo, and Bengio Yoshua.
\newblock Fitnets: Hints for thin deep nets.
\newblock {\em Proc. ICLR}, 2(3):1, 2015.

\bibitem{chen2022knowledge}
Defang Chen, Jian-Ping Mei, Hailin Zhang, Can Wang, Yan Feng, and Chun Chen.
\newblock Knowledge distillation with the reused teacher classifier.
\newblock In {\em Proceedings of the IEEE/CVF conference on computer vision and pattern recognition}, pages 11933--11942, 2022.

\bibitem{chen2017learning}
Guobin Chen, Wongun Choi, Xiang Yu, Tony Han, and Manmohan Chandraker.
\newblock Learning efficient object detection models with knowledge distillation.
\newblock {\em Advances in neural information processing systems}, 30, 2017.

\bibitem{chen2021distilling}
Pengguang Chen, Shu Liu, Hengshuang Zhao, and Jiaya Jia.
\newblock Distilling knowledge via knowledge review.
\newblock In {\em Proceedings of the IEEE/CVF Conference on Computer Vision and Pattern Recognition}, pages 5008--5017, 2021.

\bibitem{chen2021deep}
Yixin Chen, Pengguang Chen, Shu Liu, Liwei Wang, and Jiaya Jia.
\newblock Deep structured instance graph for distilling object detectors.
\newblock In {\em Proceedings of the IEEE/CVF International Conference on Computer Vision}, pages 4359--4368, 2021.

\bibitem{dai2021general}
Xing Dai, Zeren Jiang, Zhao Wu, Yiping Bao, Zhicheng Wang, Si~Liu, and Erjin Zhou.
\newblock General instance distillation for object detection.
\newblock In {\em Proceedings of the IEEE/CVF conference on computer vision and pattern recognition}, pages 7842--7851, 2021.

\bibitem{dosovitskiy2020image}
Alexey Dosovitskiy, Lucas Beyer, Alexander Kolesnikov, Dirk Weissenborn, Xiaohua Zhai, Thomas Unterthiner, Mostafa Dehghani, Matthias Minderer, Georg Heigold, Sylvain Gelly, et~al.
\newblock An image is worth 16x16 words: Transformers for image recognition at scale.
\newblock {\em arXiv preprint arXiv:2010.11929}, 2020.

\bibitem{everingham2010pascal}
Mark Everingham, Luc Van~Gool, Christopher~KI Williams, John Winn, and Andrew Zisserman.
\newblock The pascal visual object classes (voc) challenge.
\newblock {\em International journal of computer vision}, 88:303--338, 2010.

\bibitem{faster2015towards}
RCNN Faster.
\newblock Towards real-time object detection with region proposal networks.
\newblock {\em Advances in neural information processing systems}, 9199(10.5555):2969239--2969250, 2015.

\bibitem{girshick2015fast}
Ross Girshick.
\newblock Fast r-cnn in proceedings of the ieee international conference on computer vision (pp. 1440--1448).
\newblock {\em Piscataway, NJ: IEEE.[Google Scholar]}, 2, 2015.

\bibitem{girshick2014rich}
Ross Girshick, Jeff Donahue, Trevor Darrell, and Jitendra Malik.
\newblock Rich feature hierarchies for accurate object detection and semantic segmentation.
\newblock In {\em Proceedings of the IEEE conference on computer vision and pattern recognition}, pages 580--587, 2014.

\bibitem{gou2021knowledge}
Jianping Gou, Baosheng Yu, Stephen~J Maybank, and Dacheng Tao.
\newblock Knowledge distillation: A survey.
\newblock {\em International Journal of Computer Vision}, 129(6):1789--1819, 2021.

\bibitem{guo2021distilling}
Jianyuan Guo, Kai Han, Yunhe Wang, Han Wu, Xinghao Chen, Chunjing Xu, and Chang Xu.
\newblock Distilling object detectors via decoupled features.
\newblock In {\em Proceedings of the IEEE/CVF Conference on Computer Vision and Pattern Recognition}, pages 2154--2164, 2021.

\bibitem{he2017mask}
Kaiming He, Georgia Gkioxari, Piotr Doll{\'a}r, and Ross Girshick.
\newblock Mask r-cnn.
\newblock In {\em Proceedings of the IEEE international conference on computer vision}, pages 2961--2969, 2017.

\bibitem{heo2019knowledge}
Byeongho Heo, Minsik Lee, Sangdoo Yun, and Jin~Young Choi.
\newblock Knowledge transfer via distillation of activation boundaries formed by hidden neurons.
\newblock In {\em Proceedings of the AAAI conference on artificial intelligence}, volume~33, pages 3779--3787, 2019.

\bibitem{hinton2015distilling}
Geoffrey Hinton, Oriol Vinyals, and Jeff Dean.
\newblock Distilling the knowledge in a neural network.
\newblock {\em arXiv preprint arXiv:1503.02531}, 2015.

\bibitem{huang2022masked}
Tao Huang, Yuan Zhang, Shan You, Fei Wang, Chen Qian, Jian Cao, and Chang Xu.
\newblock Masked distillation with receptive tokens.
\newblock {\em arXiv preprint arXiv:2205.14589}, 2022.

\bibitem{jiao2019tinybert}
Xiaoqi Jiao, Yichun Yin, Lifeng Shang, Xin Jiang, Xiao Chen, Linlin Li, Fang Wang, and Qun Liu.
\newblock Tinybert: Distilling bert for natural language understanding.
\newblock {\em arXiv preprint arXiv:1909.10351}, 2019.

\bibitem{jin2023multi}
Ying Jin, Jiaqi Wang, and Dahua Lin.
\newblock Multi-level logit distillation.
\newblock In {\em Proceedings of the IEEE/CVF Conference on Computer Vision and Pattern Recognition}, pages 24276--24285, 2023.

\bibitem{kim2021feature}
Jangho Kim, Minsung Hyun, Inseop Chung, and Nojun Kwak.
\newblock Feature fusion for online mutual knowledge distillation.
\newblock In {\em 2020 25th International Conference on Pattern Recognition (ICPR)}, pages 4619--4625. IEEE, 2021.

\bibitem{krizhevsky2009learning}
Alex Krizhevsky, Geoffrey Hinton, et~al.
\newblock Learning multiple layers of features from tiny images.
\newblock 2009.

\bibitem{krizhevsky2012imagenet}
Alex Krizhevsky, Ilya Sutskever, and Geoffrey~E Hinton.
\newblock Imagenet classification with deep convolutional neural networks.
\newblock {\em Advances in neural information processing systems}, 25, 2012.

\bibitem{li2017mimicking}
Quanquan Li, Shengying Jin, and Junjie Yan.
\newblock Mimicking very efficient network for object detection.
\newblock In {\em Proceedings of the ieee conference on computer vision and pattern recognition}, pages 6356--6364, 2017.

\bibitem{li2020generalized}
Xiang Li, Wenhai Wang, Lijun Wu, Shuo Chen, Xiaolin Hu, Jun Li, Jinhui Tang, and Jian Yang.
\newblock Generalized focal loss: Learning qualified and distributed bounding boxes for dense object detection.
\newblock {\em Advances in Neural Information Processing Systems}, 33:21002--21012, 2020.

\bibitem{lin2017feature}
Tsung-Yi Lin, Piotr Doll{\'a}r, Ross Girshick, Kaiming He, Bharath Hariharan, and Serge Belongie.
\newblock Feature pyramid networks for object detection.
\newblock In {\em Proceedings of the IEEE conference on computer vision and pattern recognition}, pages 2117--2125, 2017.

\bibitem{lin2017focal}
Tsung-Yi Lin, Priya Goyal, Ross Girshick, Kaiming He, and Piotr Doll{\'a}r.
\newblock Focal loss for dense object detection.
\newblock In {\em Proceedings of the IEEE international conference on computer vision}, pages 2980--2988, 2017.

\bibitem{lin2014microsoft}
Tsung-Yi Lin, Michael Maire, Serge Belongie, James Hays, Pietro Perona, Deva Ramanan, Piotr Doll{\'a}r, and C~Lawrence Zitnick.
\newblock Microsoft coco: Common objects in context.
\newblock In {\em Computer Vision--ECCV 2014: 13th European Conference, Zurich, Switzerland, September 6-12, 2014, Proceedings, Part V 13}, pages 740--755. Springer, 2014.

\bibitem{liu2021exploring}
Li~Liu, Qingle Huang, Sihao Lin, Hongwei Xie, Bing Wang, Xiaojun Chang, and Xiaodan Liang.
\newblock Exploring inter-channel correlation for diversity-preserved knowledge distillation.
\newblock In {\em Proceedings of the IEEE/CVF International Conference on Computer Vision}, pages 8271--8280, 2021.

\bibitem{liu2021survey}
Yang Liu, Peng Sun, Nickolas Wergeles, and Yi~Shang.
\newblock A survey and performance evaluation of deep learning methods for small object detection.
\newblock {\em Expert Systems with Applications}, 172:114602, 2021.

\bibitem{liu2019structured}
Yifan Liu, Ke~Chen, Chris Liu, Zengchang Qin, Zhenbo Luo, and Jingdong Wang.
\newblock Structured knowledge distillation for semantic segmentation.
\newblock In {\em Proceedings of the IEEE/CVF conference on computer vision and pattern recognition}, pages 2604--2613, 2019.

\bibitem{liu2019knowledge}
Yufan Liu, Jiajiong Cao, Bing Li, Chunfeng Yuan, Weiming Hu, Yangxi Li, and Yunqiang Duan.
\newblock Knowledge distillation via instance relationship graph.
\newblock In {\em Proceedings of the IEEE/CVF Conference on Computer Vision and Pattern Recognition}, pages 7096--7104, 2019.

\bibitem{luo2003spectral}
Bin Luo, Richard~C Wilson, and Edwin~R Hancock.
\newblock Spectral embedding of graphs.
\newblock {\em Pattern recognition}, 36(10):2213--2230, 2003.

\bibitem{maroto2022benefits}
Javier Maroto, Guillermo Ortiz-Jim{\'e}nez, and Pascal Frossard.
\newblock On the benefits of knowledge distillation for adversarial robustness.
\newblock {\em arXiv preprint arXiv:2203.07159}, 2022.

\bibitem{oki2020triplet}
Hideki Oki, Motoshi Abe, Jyunichi Miyao, and Takio Kurita.
\newblock Triplet loss for knowledge distillation.
\newblock In {\em 2020 International Joint Conference on Neural Networks (IJCNN)}, pages 1--7. IEEE, 2020.

\bibitem{pang2019libra}
Jiangmiao Pang, Kai Chen, Jianping Shi, Huajun Feng, Wanli Ouyang, and Dahua Lin.
\newblock Libra r-cnn: Towards balanced learning for object detection.
\newblock In {\em Proceedings of the IEEE/CVF conference on computer vision and pattern recognition}, pages 821--830, 2019.

\bibitem{qin2021efficient}
Dian Qin, Jia-Jun Bu, Zhe Liu, Xin Shen, Sheng Zhou, Jing-Jun Gu, Zhi-Hua Wang, Lei Wu, and Hui-Fen Dai.
\newblock Efficient medical image segmentation based on knowledge distillation.
\newblock {\em IEEE Transactions on Medical Imaging}, 40(12):3820--3831, 2021.

\bibitem{shao2021and}
Rulin Shao, Jinfeng Yi, Pin-Yu Chen, and Cho-Jui Hsieh.
\newblock How and when adversarial robustness transfers in knowledge distillation?
\newblock {\em arXiv preprint arXiv:2110.12072}, 2021.

\bibitem{sun2020distilling}
Ruoyu Sun, Fuhui Tang, Xiaopeng Zhang, Hongkai Xiong, and Qi~Tian.
\newblock Distilling object detectors with task adaptive regularization.
\newblock {\em arXiv preprint arXiv:2006.13108}, 2020.

\bibitem{tian2019contrastive}
Yonglong Tian, Dilip Krishnan, and Phillip Isola.
\newblock Contrastive representation distillation.
\newblock {\em arXiv preprint arXiv:1910.10699}, 2019.

\bibitem{tian2022fully}
Zhi Tian, Xiangxiang Chu, Xiaoming Wang, Xiaolin Wei, and Chunhua Shen.
\newblock Fully convolutional one-stage 3d object detection on lidar range images.
\newblock {\em Advances in Neural Information Processing Systems}, 35:34899--34911, 2022.

\bibitem{tung2019similarity}
Frederick Tung and Greg Mori.
\newblock Similarity-preserving knowledge distillation.
\newblock In {\em Proceedings of the IEEE/CVF international conference on computer vision}, pages 1365--1374, 2019.

\bibitem{wang2022head}
Luting Wang, Xiaojie Li, Yue Liao, Zeren Jiang, Jianlong Wu, Fei Wang, Chen Qian, and Si~Liu.
\newblock Head: Hetero-assists distillation for heterogeneous object detectors.
\newblock In {\em European Conference on Computer Vision}, pages 314--331. Springer, 2022.

\bibitem{wang2019distilling}
Tao Wang, Li~Yuan, Xiaopeng Zhang, and Jiashi Feng.
\newblock Distilling object detectors with fine-grained feature imitation.
\newblock In {\em Proceedings of the IEEE/CVF Conference on Computer Vision and Pattern Recognition}, pages 4933--4942, 2019.

\bibitem{wen2021preparing}
Tiancheng Wen, Shenqi Lai, and Xueming Qian.
\newblock Preparing lessons: Improve knowledge distillation with better supervision.
\newblock {\em Neurocomputing}, 454:25--33, 2021.

\bibitem{wu2024regional}
Pingfan Wu, Jiayu Zhang, Han Sun, and Ningzhong Liu.
\newblock Regional filtering distillation for object detection.
\newblock {\em Machine Vision and Applications}, 35(2):24, 2024.

\bibitem{yang2023context}
Aijia Yang, Sihao Lin, Chung-Hsing Yeh, Minglei Shu, Yi~Yang, and Xiaojun Chang.
\newblock Context matters: Distilling knowledge graph for enhanced object detection.
\newblock {\em IEEE Transactions on Multimedia}, 2023.

\bibitem{yang2022cross}
Chuanguang Yang, Helong Zhou, Zhulin An, Xue Jiang, Yongjun Xu, and Qian Zhang.
\newblock Cross-image relational knowledge distillation for semantic segmentation.
\newblock In {\em Proceedings of the IEEE/CVF Conference on Computer Vision and Pattern Recognition}, pages 12319--12328, 2022.

\bibitem{yang2022focal}
Zhendong Yang, Zhe Li, Xiaohu Jiang, Yuan Gong, Zehuan Yuan, Danpei Zhao, and Chun Yuan.
\newblock Focal and global knowledge distillation for detectors.
\newblock In {\em Proceedings of the IEEE/CVF Conference on Computer Vision and Pattern Recognition}, pages 4643--4652, 2022.

\bibitem{yim2017gift}
Junho Yim, Donggyu Joo, Jihoon Bae, and Junmo Kim.
\newblock A gift from knowledge distillation: Fast optimization, network minimization and transfer learning.
\newblock In {\em Proceedings of the IEEE conference on computer vision and pattern recognition}, pages 4133--4141, 2017.

\bibitem{zagoruyko2016paying}
Sergey Zagoruyko and Nikos Komodakis.
\newblock Paying more attention to attention: Improving the performance of convolutional neural networks via attention transfer.
\newblock {\em arXiv preprint arXiv:1612.03928}, 2016.

\bibitem{zhang2023structured}
Linfeng Zhang and Kaisheng Ma.
\newblock Structured knowledge distillation for accurate and efficient object detection.
\newblock {\em IEEE Transactions on Pattern Analysis and Machine Intelligence}, 2023.

\bibitem{zheng2023localization}
Zhaohui Zheng, Rongguang Ye, Qibin Hou, Dongwei Ren, Ping Wang, Wangmeng Zuo, and Ming-Ming Cheng.
\newblock Localization distillation for object detection.
\newblock {\em IEEE Transactions on Pattern Analysis and Machine Intelligence}, 2023.

\bibitem{zhixing2021distilling}
Du~Zhixing, Rui Zhang, Ming Chang, Shaoli Liu, Tianshi Chen, Yunji Chen, et~al.
\newblock Distilling object detectors with feature richness.
\newblock {\em Advances in Neural Information Processing Systems}, 34:5213--5224, 2021.

\bibitem{zhou2021distilling}
Sheng Zhou, Yucheng Wang, Defang Chen, Jiawei Chen, Xin Wang, Can Wang, and Jiajun Bu.
\newblock Distilling holistic knowledge with graph neural networks.
\newblock In {\em Proceedings of the IEEE/CVF international conference on computer vision}, pages 10387--10396, 2021.

\bibitem{zi2021revisiting}
Bojia Zi, Shihao Zhao, Xingjun Ma, and Yu-Gang Jiang.
\newblock Revisiting adversarial robustness distillation: Robust soft labels make student better.
\newblock In {\em Proceedings of the IEEE/CVF International Conference on Computer Vision}, pages 16443--16452, 2021.

\end{thebibliography}


\appendix

\section{Related Work}
\label{related work}

\subsection{Knowledge Distillation}

Knowledge distillation is a model compression technique that has been applied across various tasks and domains, including image classification \cite{hinton2015distilling},object detection \cite{zhang2023structured, wang2019distilling, chen2017learning, wu2024regional}, natural language processing \cite{jiao2019tinybert}, semantic segmentation \cite{qin2021efficient, liu2019structured}, and more. In recent years, it has also been explored for enhancing model generalization capabilities, reducing model reliance on data biases, and improving model performance on specific tasks, such as learning with limited labeled data. Hinton \cite{hinton2015distilling}and others \cite{tian2019contrastive, oki2020triplet, wen2021preparing} posited that the logit outputs of a teacher model, relative to one-hot labels, contain rich interpretive information. Minimizing the KL divergence difference between student and teacher outputs can help the student approximate the teacher's decision boundary. As research progressed, researchers discovered that the rich feature representations in the intermediate layers of the teacher model contain additional information useful for the learning task. FitNet \cite{adriana2015fitnets} proposed treating all features equally in distillation. DeFeat \cite{guo2021distilling} introduced hyperparameters to weight foreground and background features to overcome the adverse effects of uneven positive and negative regions. To match the semantic levels between teacher and student feature layers, AT \cite{zagoruyko2016paying} used attention maps as carriers of knowledge transfer and compared the distillation performance of activation-based and gradient-based attention maps. Guobin Chen \cite{chen2017learning} were pioneers in introducing knowledge distillation techniques to object detection tasks. Relational distillation goes further than feature distillation by exploring interactions and dependencies between instances. The intrinsic structure of the data, as a more abstract and general form of knowledge, aids the student in mimicking the teacher's generalization capabilities. SP \cite{tung2019similarity} aligned pair-wise similarities between teacher and student features, encoding relational knowledge into an adjacency matrix refined into the student model. Sheng Zhou et al. unified the representation of individual knowledge in Euclidean metric spaces and sample correlations as a graph network based on vertices and edges. Total mutual information was used as an additional supervisory signal to guide the training of the student model. Yufan Liu et al. \cite{liu2019knowledge}, by constructing instance relation graphs to simulate the knowledge in network layers, proposed the IRG transformation to smooth the transition between feature layers, showing strong robustness to different network architectures.

FSP \cite{yim2017gift} addresses the pairwise similarity between feature layers. Li Liu and others \cite{liu2021exploring} explored the contribution of homogeneity and diversity among channels to visual tasks, proposing the ICKD method, which is based on the correlation between feature channels. Pixel-level relational distillation, as exemplified by CIRKD, is primarily used for segmentation tasks \cite{yang2022cross}.

\subsection{Object detection}

Compared to classification, detection provides an understanding of the image's foreground and background \cite{liu2021survey}.The landscape of object detection has been predominantly shaped by deep learning techniques, leveraging Convolutional Neural Networks (CNNs) to achieve remarkable feats \cite{krizhevsky2012imagenet}. These methods are broadly categorized into two distinct paradigms: two-stage and one-stage detectors.Two-stage Detectors: Pioneered by the R-CNN family, including Fast R-CNN and Faster RCNN \cite{girshick2014rich, girshick2015fast, faster2015towards, he2017mask, pang2019libra}, this paradigm initially generates region proposals to highlight areas of interest and subsequently performs classification and bounding box regression on each proposal. One-stage anchor-based detectors, such as RetinaNet \cite{lin2017focal}, predict class scores and bounding boxes directly from the image, utilizing predefined anchor boxes to accommodate a variety of scales and aspect ratios. Meanwhile, anchor-free detectors like FCOS \cite{tian2022fully} eliminate the necessity for anchor boxes by directly predicting object key points or centers. FPN \cite{lin2017feature}  constructs a multi-level feature pyramid, effectively integrating spatial information across different scales through a top-down pathway and lateral connections, thereby achieving precise detection of objects of varying sizes.

\subsection{Knowledge Distillation Based on Graphs}

By constructing relational graphs, we provide an intuitive understanding of abstract relational knowledge distillation. Vertices on the graph represent feature information, while edges depict the interactions and dependencies between pieces of knowledge. IRG \cite{liu2019knowledge} models the relationships between instances at each layer as a graph. Yang (CM) \cite{yang2023context} utilizes a transformer to perform message passing and aggregation operations on the graph, achieving a comprehensive representation of vertices that captures both the local neighborhood structure and the global graph structure. Chen et al. \cite{chen2021deep} modeled the relationships between proposal-level instance features within a Region Proposal Network as a structured graph, conveying both local and global feature information. Holistic KD \cite{zhou2021distilling} posits that individual knowledge and relational knowledge offer mutually orthogonal feature representations. Integrating both into a unified graph-based embedding, this encoded knowledge representation is then transferred to the student network. In our method, we do not introduce any additional auxiliary modules; instead, by extracting the spectral embedding of the relational graph, we facilitate the propagation and aggregation of representations from vertices and their highly similar neighboring regions.

\section{Experiments Setting}
\label{Experiments Setting}

\textbf{MS-COCO} is a widely used large-scale object detection dataset that includes 120K training images and 5K validation images across 80 object categories and 91 scene categories.

\textbf{Pascal VOC} dataset encompasses 20 object categories, including people, animals, vehicles, and indoor objects, among others.

\textbf{CIFAR-100} dataset, designed for classification tasks, comprises 60,000 color images with a resolution of 32x32 pixels. It includes 50,000 training images and 10,000 validation images. The images are categorized into 100 classes, which are further grouped into 20 superclasses.

We test the performance of our method on various datasets to demonstrate its generalization ability.

We conducted a comprehensive performance comparison between our proposed method and other state-of-the-art (SOTA) methods on the COCO dataset. Specifically, we detailed the performance of our method with various mainstream detectors, including two-stage detectors such as Faster RCNN \cite{faster2015towards} with FPN \cite{lin2017feature}, one-stage anchor-based detectors like RetinaNet \cite{lin2017focal}, and one-stage anchor-free detectors such as FCOS \cite{tian2022fully}, as well as other detectors including GFL \cite{li2020generalized}. The evaluation metrics we established on the COCO dataset include AP, 
$\text{AP}_{50}$, $\text{AP}_{75}$, $\text{AP}_{S}$, $\text{AP}_{M}$, and $\text{AP}_{L}$. We also designed experiments involving both homogeneous and heterogeneous teacher-student network architectures, such as T-ResNet101 with S-ResNet50 (homogeneous) and T-ResNet101 with S-MobileNetV2 (heterogeneous). To demonstrate the versatility of our method, we evaluated its performance on the VOC2007 testing set, which consists of 4952 images, across three typical detection architectures: Faster RCNN \cite{faster2015towards}, RetinaNet \cite{lin2017focal}, and FCOS \cite{tian2022fully}. Specifically, we assessed the AP, $\text{AP}_{50}$, and $\text{AP}_{75}$ metrics for each architecture. For classification tasks, we utilize top-1 accuracy as the primary criterion to assess model performance. Similarly, we have conducted experiments to evaluate the classification accuracy of our method across a variety of network architectures, including ResNet, Wide Residual Networks (WRN), VGG, ShuffleNetV2, and MobileNetV2.

We employed the SGD optimizer for updating the model parameters, with momentum configured at 0.9, an initial learning rate  set to 0.01, batch size determined at 16, and weight decay fixed at 0.01. All training utilizes AMP (Automatic Mixed Precision) optimization to accelerate the training process, and for the teacher networks across all datasets, we employ a 2x schedule training plan. The training was executed on eight Tesla 32G V100 GPUs on the AutoDL cloud platform. We have summarized the computational costs  of our method when applied to various backbones within the GFL \cite{li2020generalized} framework, as detailed in Table \ref{Training cost}. Our approach utilizes the hyperparameters $\alpha$, $\beta$ and $\gamma$ to achieve a balance between graph loss and spectral embedding loss. Without prior specification, we treat all supervision signals with equal importance. Due to the constraints of paper length, not all results could be fully displayed.
\begin{table}[ht]
\centering
\caption{Training costs}
\label{Training cost}
\begin{tabular}{lccccc}
\hline
 Backbone & Training hours & Schedule & FLOP & \#Param.  \\
\hline
 ResNet101 & 14.3h & 2x & 297.6G & 51.1M  \\
 MobileNetV2 & 10.7h & 1x & 130.2G & 10.4M  \\
 ResNet18 & 8.3h & 1x & 162.8G & 19.1M  \\
 ResNet34 & 9.1h & 1x & 202.7G & 29.0M  \\
ResNet50 & 10.5h & 1x & 214.7G & 32.1M  \\
\hline
\end{tabular}
\end{table}

\section{More Experiments}
\label{More Experiments}

\subsection{Pascal VOC}
\label{PASCAL VOC}

As demonstrated in Table \ref{voc}, our experiments on the VOC2007 test set revealed the performance of our method across three typical detection architectures (Faster RCNN \cite{faster2015towards}, RetinaNet \cite{lin2017focal}, and FCOS \cite{tian2022fully}). We maintained ResNet101 as the guiding teacher for the student ResNet50. Our method brings notable AP gains, with student models generally outperforming teacher models. For example, in the Faster RCNN \cite{faster2015towards} setting, the distilled student model achieves a lead of 1.9 AP over the teacher network. In the case of lightweight one-stage detectors, our method particularly enhances localization accuracy. For instance, the student model of RetinaNet \cite{lin2017focal} exhibits an 8.7\% improvement in $\text{AP}_{75}$, surpassing GI \cite{dai2021general} by 4.6\%. We visualized the attention distribution of RetinaNet \cite{lin2017focal} in the form of a heatmap. As shown in Figure~\ref{figure 6}, we have compared the results of SE, GI \cite{dai2021general}, students, and teachers side by side. Our approach is most closely aligned with that of the teacher.

\begin{table}[ht]
\caption{Performance on VOC. FGFI is only applicable to two-stage and anchor-based scenarios.}
\label{voc}
\centering
\resizebox{\textwidth}{!}{
\begin{tabular}{l|p{0.9cm}ccp{0.8cm}ccp{0.7cm}cc}
\hline
\multirow{2}{*}{Method} & \multicolumn{3}{c|}{Faster RCNN \cite{faster2015towards} Res101-50} & \multicolumn{3}{c|}{RetinaNet \cite{lin2017focal} Res101-50}& \multicolumn{3}{c}{FCOS \cite{tian2022fully} Res101-50} \\  \cline{2-10}
 & AP &AP$_{50}$&\multicolumn{1}{c|}{AP$_{75}$} & AP & AP$_{50}$ &\multicolumn{1}{c|}{AP$_{75}$}& AP & AP$_{50}$ &AP$_{75}$ \\
 \hline
Teacher&56.3&82.7& \multicolumn{1}{c|}{62.6}&58.2&82.0&\multicolumn{1}{c|}{63.0}&58.4&81.6&64.3\\
Student&54.2&82.1& \multicolumn{1}{c|}{59.9}&56.1&80.9&\multicolumn{1}{c|}{60.7}&56.1&80.0&61.6\\
FitNet \cite{adriana2015fitnets}&55.0&82.2& \multicolumn{1}{c|}{61.2}&56.4&81.7&\multicolumn{1}{c|}{61.7}&57.0&80.3&62.1\\
*FGFI \cite{wang2019distilling}&55.3&82.1& \multicolumn{1}{c|}{61.1}&55.6&81.4&\multicolumn{1}{c|}{60.5}& \textemdash&\textemdash&\textemdash\\
GI \cite{dai2021general}&56.5&82.6& \multicolumn{1}{c|}{61.6}&57.9&82.0&\multicolumn{1}{c|}{63.2}&58.4&81.3&62.9\\
\rowcolor{gray!25}
\textbf{Ours}&\textbf{58.2}&\textbf{83.6}&\multicolumn{1}{c|}{\textbf{63.9} }&\textbf{59.9}&\textbf{83.1}&\multicolumn{1}{c|}{\textbf{65.8}}&\textbf{59.5}&\textbf{83.2}&\textbf{65.6}\\
\hline
\end{tabular}
}
\end{table}

\subsection{Attention Mechanism Ablation Study}
\label{Attention Mechanism Ablation}

\begin{wraptable}{r}{0.5\textwidth} 
  \caption{Conducting ablation experiments on attention mechanisms with the Faster RCNN-FPN detector in a ResNet101-50 teacher-student architecture on the COCO dataset.}
  \label{table 8}
  \centering
  \begin{tabular}{cc|ccc}
    \toprule
    \multicolumn{2}{c|}{\textbf{Attention}} & \multicolumn{3}{c}{\textbf{Result}} \\
    \midrule
    Vertex & Edge & AP & AP$_{50}$ & AP$_{75}$ \\
    \midrule
    \ding{55} & \ding{55} & 40.9 & 60.7 & 43.8 \\
    \checkmark & \ding{55} & 41.4 & 61.6 & 45.1 \\
    \ding{55} & \checkmark & 41.2 & 60.9 & 44.7 \\
    \rowcolor{gray!25}
    \checkmark & \checkmark & \textbf{41.9} & \textbf{61.8} & \textbf{45.5} \\
    \bottomrule
  \end{tabular}
\end{wraptable}

As shown in the Table \ref{table 8}. When only vertex attention is enabled, the model sees an improvement across all three metrics: AP increases to 41.4, AP$_{50}$ rises to 61.6, and AP$_{75}$ grows to 45.1. With just the edge attention activated, there's a slight bump in AP to 41.2, AP$_{50}$ remains competitive at 60.9, and there's a more noticeable boost in AP$_{75}$, reaching 44.7. However, when both attention mechanisms are employed, the model achieves the best outcomes on all metrics: AP further advances to 41.9, AP$_{50}$ climbs to 61.8, and AP$_{75}$ ascends to 45.5. The integration of both vertex and edge attention mechanisms contributes to the most substantial performance gains, underscoring their complementary roles in enhancing the model's capability to discern and process spatial relationships within the data.

\subsection{Sensitivity Analysis of the Hyperparameter N}
\label{N}

\begin{figure}[htbp]
  \centering
  \begin{minipage}{0.48\textwidth}
    \centering
    \includegraphics[width=0.95\linewidth]{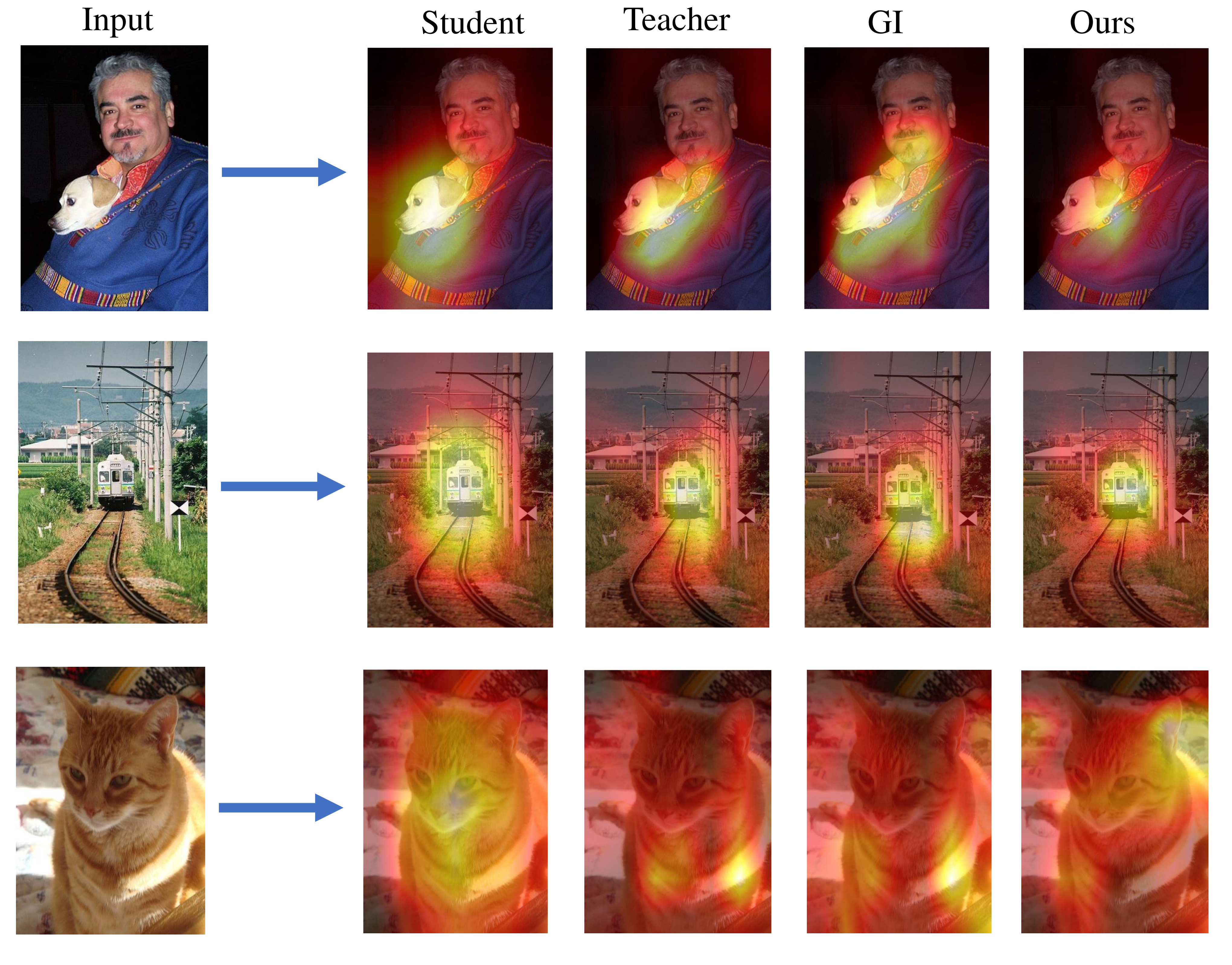}
    \caption{The attention distribution of different types of detectors.}
    \label{figure 6}
  \end{minipage}\hfill
  \begin{minipage}{0.48\textwidth}
    \centering
    \includegraphics[width=1\linewidth]{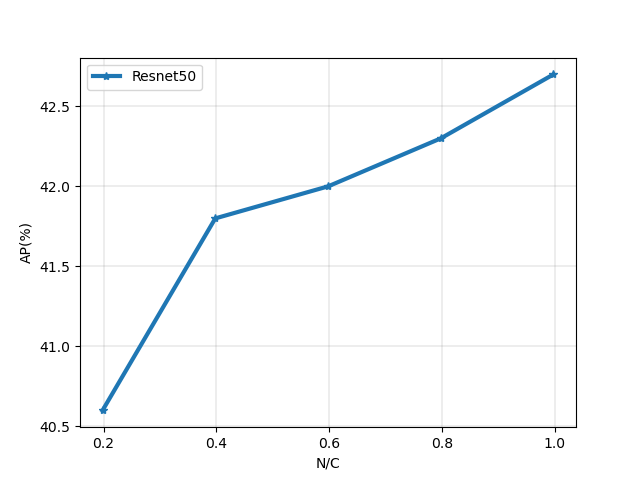}
    \caption{Sensitivity experiment curve of the hyperparameter N}
    \label{figure 7}
  \end{minipage}
\end{figure}

In our methodology, the chosen detector is GFL, and the teacher network employed is ResNet101. 
In spectral embedding, the number of eigenvectors of the Laplacian matrix can reach its dimensionality, corresponding to the number of vertices in the graph. This is closely related to the spectral properties of the Laplacian, encoding the graph's topology. As illustrated in Figure \ref{figure 7}, the selection of eigenvectors (denoted as N/C, where N is the number of selected eigenvectors and C is the number of channels) affects the performance metric AP\%, which stands for Average Precision Percentage. The graph demonstrates how increasing the proportion of utilized eigenvectors relative to the number of channels enhances the discriminative power of the resulting spectral features.


\section{Limitations}
\label{Limitations}

One inherent limitation of our framework pertains to the subjective nature of feature and channel selection for distillation. Our method necessitates identifying specific features and channels that are deemed crucial for effectively transferring knowledge from the teacher to the student model. This selection process is inherently influenced by the criteria and strategies we employ, which introduces a degree of subjectivity.


\end{document}